\documentclass[sn-mathphys,Numbered]{sn-jnl}
\usepackage{graphicx}%
\usepackage{multirow}%
\usepackage{amsmath,amssymb,amsfonts}%
\usepackage{amsthm}%
\usepackage{mathrsfs}%
\usepackage[title]{appendix}%
\usepackage{xcolor}%
\usepackage{textcomp}%
\usepackage{manyfoot}%
\usepackage{booktabs}%
\usepackage{algorithm}%
\usepackage{algorithmicx}%
\usepackage{algpseudocode}%
\usepackage{listings}%
\usepackage{multirow}
\usepackage{makecell}
\usepackage{amsmath}
\usepackage{rotating}

\begin{document}

\title[CoFiNet]{CoFiNet: Unveiling Camouflaged Objects with Multi-Scale Finesse}


\author[1,2]{\fnm{Cunhan} \sur{Guo}}\email{guocunhan22@mails.ucas.ac.cn} 

\author*[1,3]{\fnm{Heyan} \sur{Huang}}\email{hhy63@bit.edu.cn}

\affil*[1]{\orgdiv{School of Emergency Management Science and Engineering}, \orgname{University of Chinese Academy of Sciences}, \orgaddress{\street{No.1, Yanqihu East Road}, \city{Beijing}, \postcode{101400}, \country{China}}}

\affil[2]{\orgdiv{Southeast Academy of information Technology}, \orgname{Beijing Institute of Technology}, \orgaddress{\street{1998 Licheng Middle Avenue}, \city{Putian}, \postcode{351100}, \country{China}}}

\affil[3]{\orgdiv{School of Computer Science and Technology}, \orgname{Beijing Institute of Technology}, \orgaddress{\street{No.5 Zhongguancun South Street}, \city{Beijing}, \postcode{100086}, \country{China}}}

\abstract{Camouflaged Object Detection (COD) is a critical aspect of computer vision aimed at identifying concealed objects, with applications spanning military, industrial, medical and monitoring domains. To address the problem of poor detail segmentation effect, we introduce a novel method for camouflage object detection, named CoFiNet. Our approach primarily focuses on multi-scale feature fusion and extraction, with special attention to the model's segmentation effectiveness for detailed features, enhancing its ability to effectively detect camouflaged objects. CoFiNet adopts a coarse-to-fine strategy. A multi-scale feature integration module is laveraged to enhance the model's capability of fusing context feature. A multi-activation selective kernel module is leveraged to grant the model the ability to autonomously alter its receptive field, enabling it to selectively choose an appropriate receptive field for camouflaged objects of different sizes. During mask generation, we employ the dual-mask strategy for image segmentation, separating the reconstruction of coarse and fine masks, which significantly enhances the model's learning capacity for details. Comprehensive experiments were conducted on four different datasets, demonstrating that CoFiNet achieves state-of-the-art performance across all datasets. The experiment results of CoFiNet underscore its effectiveness in camouflage object detection and highlight its potential in various practical application scenarios.
}




\maketitle

\section{Introduction}\label{sec1}
In the natural world, numerous animals possess remarkable camouflage abilities, such as chameleons and stick insects. They achieve self-protection by altering their colors or mimicking their surroundings. The detection of such objects is referred to as narrow-sense camouflage object detection. In a broader sense, camouflage object detection extends to the detection of low-detectability objects, including those challenging to identify through traditional methods due to factors like color, texture, or shape. This field's definition encompasses identifying potential threats in military applications, as well as meeting the detection needs of concealed targets in industries, medicine, monitoring, protection, and unmanned driving. For instance, the detection of lung nodules smaller than 5 millimeters in size is crucial in intelligent analysis of Computed Tomography images. These nodules are highly prone to being overlooked and require accurate detection and segmentation to aid doctors in better diagnosing and treating patients. This exemplifies a significant practical application scenario for camouflage object detection.

Currently, there are two main approaches to camouflage object detection. The first relies on purely visual methods, utilizing optical imaging to directly detect and segment camouflage objects based on subtle differences between the object and the background \cite{early-1,early-2,early-3,early-4}. In earlier years, researchers often employed hand-crafted features to extract key information about the edges and contours of camouflage objects, enabling segmentation. While this method is computationally fast, it heavily depends on manually designed features and exhibits poor generalization. With the advancement of deep learning technology, contemporary approaches increasingly utilize deep neural networks for feature extraction, achieving end-to-end segmentation and significantly improving algorithm generalization by decoupling reliance on handcrafted features. The second approach involves multi-modal information fusion. Due to the limitations of traditional optical sensors in detecting camouflage objects, researchers have gradually introduced infrared imagery \cite{infrared}, hyperspectral imaging \cite{hyperspectral} , and polarized imaging \cite{polarization} as supplements to enhance detection accuracy.

Currently, camouflage object detection still faces several formidable challenges \cite{survey}. The fundamental challenge lies in the high similarity between camouflage objects and their backgrounds. Camouflage objects mimic the colors, textures, and brightness of their surroundings, making them difficult to distinguish from the background in images. This similarity poses a difficulty for models to discern subtle differences between camouflage objects and the background, requiring sensitivity to fine features. Additionally, camouflage objects exhibit various scales, implying that they may appear in different sizes in images. Due to factors like distance, angle, or other variables, the scale of camouflaged targets may vary, necessitating detection models to adapt flexibly to multi-scale camouflage objects, ensuring accurate detection at different observation distances or angles. Finally, camouflage objects may experience partial occlusion, rendering their representation in the image incomplete. This occlusion can arise from the surrounding environment or other objects, increasing the difficulty for the model to recognize the edges and shapes of camouflaged objects. Therefore, models need to possess the capability to handle partially occluded situations effectively, addressing the incomplete presentation of camouflaged targets in images. In addition to the inherent challenges posed by camouflage, camouflage object detection tasks also encounter limitations from deep learning networks. General deep learning object detection networks face various limitations. The similarity in color and texture makes it challenging for deep learning models to differentiate camouflage objects from the background. Issues such as irregular shapes, scale variations, and insufficient utilization of keypoint information further contribute to the difficulty of detection.

To adress these challenges, we propose the Coarse to Fine Network (CoFiNet), which adopts a multi-scale and multi-region strategy, progressively refining feature extraction and mask generation. To facilitate the fusion of multi-scale features, we design the Multi-Scale Feature Integration module (MSFI), ensuring comprehensive integration of data from various scales. For feature extraction, we introduce the Multi-activation Selective Kernel Module(MSKM), providing the model with the ability to autonomously alter its receptive field. This enables the model to selectively choose an appropriate receptive field for camouflaged objects of different sizes, achieving the dual purpose of extracting both coarse and fine-grained features.  During mask generation, we utilize the dual-mask strategy for image segmentation, which separates the reconstruction of coarse and fine masks. The final features are then fused, placing a significant emphasis on enhancing the model's learning capacity for intricate details.

The contributions of this article can be summarized as follows:

1. We proposed a powerful framework, the CoFiNet, which enhances the detection and segmentation of camouflaged objects by transitioning from a global to local perspective and from coarse to fine granularity. This significantly improves the model's ability to segment intricate details of camouflaged objects.

2. The multi-scale feature integration module  and multi-activation selective kernel module effectively enhance the model's capability to fuse and extract features of different granularities and scales.

3. We introduced a dual-mask strategy for image segmentation, which, building upon the model's coarse segmentation, further detects and segments details, thereby enhancing the overall performance of CoFiNet.

4. Comprehensive experimentation with the proposed CoFiNet on four datasets, yielding state-of-the-art (SOTA) results across all datasets, showcasing the effectiveness of our proposed approach.

The remaining sections of the article are organized as follows: Section 2 provides a comprehensive review of related works. In Section 3, we delve into the detailed presentation of our model's methodology. Section 4 validates the effectiveness of our algorithm through extensive experimentations, Section 5 concludes the entire article.

\section{Related Work}
\subsection{Obeject detection}
Obeject detection, as a prominent avenue within the realm of computer vision, has witnessed substantial advancements in recent years. It can be classified into three distinct categories: General Object Detection (GOD), Salient Object Detection (SOD) \cite{sod}, and Camouflage Object Detection \cite{sinet,sinetv2}. In the subsequent sections, each of these three categories will be meticulously reviewed.
\subsubsection{Generic object detection}
The evolution of the generic object detection field over the past two decades can be divided into two distinct historical periods: the traditional object detection era (1990s-2014) and the deep learning-based detection era (after 2014) \cite{survey}. During the traditional period, algorithms primarily relied on manually designed features and acceleration techniques, with representative methods including the Viola-Jones detector \cite{viola}, HOG detector \cite{hog}, and Deformable Part Models \cite{dpm}. With the rise of deep learning, detection methods based on deep learning have flourished. Detectors such as the RCNN series \cite{rcnn} and the Feature Pyramid Network \cite{fpn} introduced techniques such as region proposal networks and multi-scale detection, achieving end-to-end learning frameworks and higher performance. In the new era, single-stage detectors like YOLO \cite{yolo}, SSD \cite{ssd}, and RetinaNet \cite{retinaNet} adopt a single-stage detection paradigm, balancing speed and accuracy.

In terms of technological evolution, multi-scale detection has transitioned from the feature pyramid and sliding window approach to multi-reference or multi-resolution detection, gradually realizing more direct and effective generic object detection. Context initiation techniques have evolved from local context detection to global context detection and, more recently, to immediate context interaction, progressively considering the dependencies between objects and their surroundings. Hard Negative Mining (HNM) techniques have made progress in addressing training imbalance issues, evolving from early bootstrapping methods to HNM techniques in the deep learning era. By reintroducing bootstrapping methods and designing new loss functions, these techniques effectively cope with data imbalance. The evolution of loss functions has incorporated methods such as label smoothing and focal loss, enhancing classification efficiency and addressing issues related to class imbalance and difficulty differences in classification. Non-Maximum Suppression (NMS) techniques have evolved from early greedy selection to learning-based NMS, and further to detectors without NMS, representing a potential direction for future generic object detection.

\subsubsection{Salient object detection}
As a crucial step in image preprocessing, salient object detection finds widespread applications across various computer vision tasks. It is employed for guiding image description by utilizing salient content \cite{sod_baseline}, transforming unsupervised learning into multi-instance learning for localization and classification tasks \cite{usl_sod}, aiding unsupervised video object segmentation with object-level clues \cite{seg_sod}, providing saliency contour information for object detection tasks \cite{detect_sod}, and constructing datasets to evaluate the performance of visual question-answering models \cite{supercnn}, among other applications.

Salient object detection can be categorized into sparse detection methods and dense detection methods. Sparse detection methods further divide into superpixel methods and candidate object methods based on different processing units. Superpixel methods segment images into multiple superpixels of varying sizes at different scales, compute saliency maps, and finally, merge the results from different scales to form salient object regions. Candidate object methods generate a large number of candidate objects, predict the saliency map for each candidate object, and sum them to obtain the salient object detection result for the entire image.

Dense detection methods primarily revolve around encoder-decoder architectures, with U-Net-based networks being the most commonly used. U-Net is favored for its ability to address issues where the model lacks low-level spatial details, leading to saliency detection results with poorly-maintained boundaries. Current major improvement methods include enhancements to skip connections between the encoder and decoder, the introduction of recursive structures, methods for integrating contextual semantic information, and techniques for reinforcing the detection of salient object boundaries.

\subsubsection{Camouflaged object detection}
In recent years, significant progress has been made in camouflaged target detection using deep learning, leading to the emergence of various algorithms that improve detection performance. Here, we analyze these methods in detail from different strategies:

\textbf{Coarse-to-fine strategy:}
Typical representatives such as SuperCNN generate superpixels using the SLIC algorithm and combine convolutional neural networks to fuse hierarchical features, ultimately generating a saliency map for target detection \cite{supercnn}. MCDL focuses on extracting contextual information, with each superpixel as the center, using two windows to extract local and global contextual information to improve salient target detection. ELD introduces manually extracted salient features from traditional methods and combines them with MCDL to provide complementary information to the deep convolutional network, thereby improving salient detection performance.

\textbf{Multi-task learning strategy:}
Multi-task learning enhances camouflaged target detection performance by introducing auxiliary tasks such as classification, localization, and binary segmentation. These methods include classification + segmentation, localization/sorting + segmentation, mimicry attack + segmentation, texture detection + segmentation, and edge detection + segmentation. By working collaboratively, these multi-task methods can extract more comprehensive camouflaged target information and improve detection performance \cite{multitask,nc4k,multitask2,multitask3,multitask4,multitask5}.

\textbf{Confidence-aware learning strategy:}
The confidence-aware learning aims to estimate uncertainty representing data quality or perceived uncertainty of the true model \cite{confidence}. This includes strategies such as adversarial training, dynamic supervision, and regularization constraints to improve the robustness of camouflaged target detection.

\textbf{Multi-source information fusion strategy:}
Multi-source information fusion enhances camouflaged target detection performance by incorporating depth information, frequency domain information, and other supplementary information to RGB information \cite{infrared, polarization, hyperspectral}. Methods include RGB-D fusion and frequency domain information fusion, but they also face challenges such as inaccurate depth information and fusion complexity.

\textbf{Camouflaged target detection based on Transformer:}
The Transformer has been introduced into camouflaged target detection to capture long-range dependencies using the self-attention mechanism and improve the model's ability to capture global information \cite{transformer}. Different Transformer models such as ViT, PVT, SETR, and others have been applied in camouflaged target detection tasks, effectively improving model performance.

\subsection{Multi-scale feature fusion}
In the field of computer vision, multi-scale feature fusion is a crucial concept aimed at effectively handling feature information at different scales within images. Relevant research in this domain encompasses various aspects, providing a more comprehensive and efficient feature representation for image processing and computer vision tasks.

Firstly, concerning the fusion of features with a pyramid structure, a series of studies are dedicated to constructing pyramid structures to capture features at multiple scales. By extracting features at different levels of the pyramid and merging them, researchers have successfully achieved a more comprehensive acquisition of image information. This approach demonstrates significant advantages in handling complex scenes with features at different scales, such as natural landscapes or medical images. Secondly, cross-scale connections in Convolutional Neural Networks have garnered considerable attention. Some works focus on fusing features from different levels through mechanisms like skip connections or attention mechanisms. This approach excels in improving the model's recognition and localization capabilities, particularly when dealing with complex image structures. In the research on multi-scale feature fusion, the application of attention mechanisms has become increasingly widespread. By allocating different attention weights at different scales, models can more effectively handle various details and structures, enhancing their understanding and expressive capabilities for image content. The introduction of methods such as Graph Convolutional Networks  provides new pathways for multi-scale information propagation. These networks introduce graph structures into feature fusion, better capturing relationships between pixels in images. The integration of multi-scale information through graph-based propagation further enhances the model's performance in complex scenes. Lastly, research on adaptive feature fusion explores methods to adaptively adjust multi-scale feature fusion based on the content of input images. This flexible fusion approach aims to improve the model's adaptability to different scenes and objects, making it more robust and capable of generalization.

In summary, these research efforts collectively form the domain of multi-scale feature fusion, providing significant impetus for the development of computer vision. With the continuous development and innovation of these methods, we anticipate witnessing further advancements in computer vision's ability to handle complex image tasks.

\section{Methodology}
\subsection{Overview of CoFiNet}
\begin{figure}[htbp]
\centering
\includegraphics[width=\textwidth]{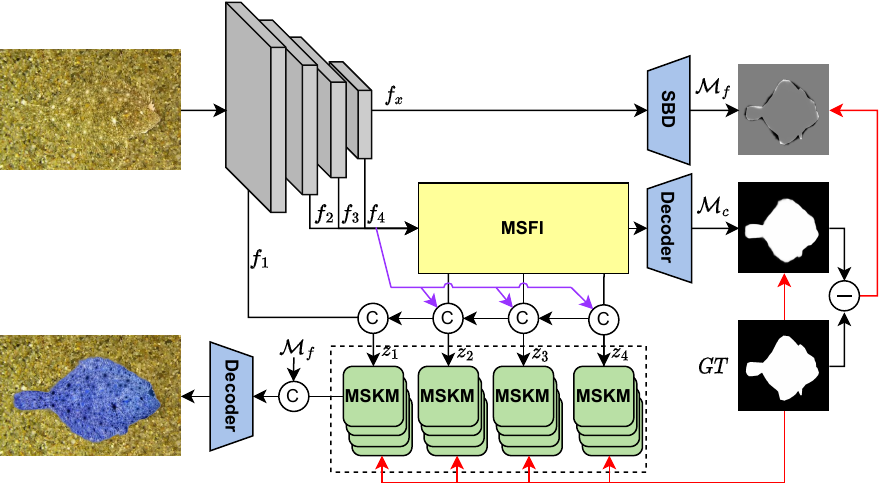}
\caption{\centering{Overview of CoFiNet, where the red arrows stand for the deep supervision.}}
\label{fig:main}
\end{figure}

As illustrated in \autoref{fig:main}, our model adopts an overall deep supervised U-Net architecture. The overall model can be divided into four stages: the encoding stage, the feature fusion and integration stage, the feature extraction stage, and the feature decoding stage. 

After inputting the image, we utilize a large-scale pretrained model (such as SwinV2 \cite{swinv2} or PVTv2 \cite{pvtv2}) to obtain pyramid features $f_i$, where $i \in \{1, 2, 3, 4\}$ and the final feature $f_x \in \mathbb{R}^{(B,L)}$. The features $f_2, f_3, f_4$ are fed into MSFI for multi-scale feature fusion, producing $f_2'', f_3''$ and $ f_4''$, which are then input into MSKM for more detailed feature extraction. It is worth noting that, to guide MSFI in preserving spatial features better, we input $f_2''$ into the decoder, supervise the decoding results, and obtain the coarse mask denoted as $\mathcal{M}_c$. To enhance the model's ability to capture details, we subtract the Ground Truth (GT) from $\mathcal{M}_c$ to obtain detail feature labels that are challenging for the model to capture. Further, we use the Spatial Broadcast Decoder (SBD) to model these details and obtain the fine mask $\mathcal{M}_f$. During final decoding, $\mathcal{M}_f$ is concatenated with the output of MSKM, serving as joint input to the decoder, thereby generating the detection results for camouflaged objects. In the following subsections, detailed explanations will be provided for each of the mentioned modules.

\subsection{Multi-scale feature integration module}
\begin{figure*}[htbp]
\centering
\includegraphics[width=0.5\textwidth]{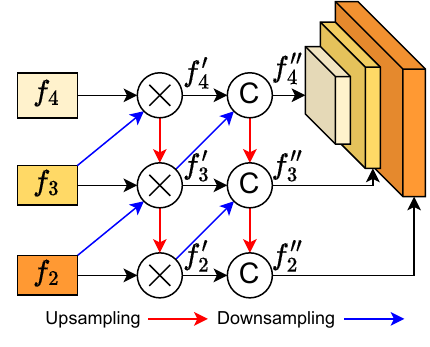}
\caption{\centering{Multi-scale feature integration module}}
\label{fig:msfi}
\end{figure*}

After obtaining multi-scale features using the encoder, we designed a multi-scale feature integration module, as illustrated in the \autoref{fig:msfi}, to fuse these features. For two reasons, we do not integrate the output of the first stage of the encoder. First, the size of $f_1$ is large, requiring significant computational resources for fusion. Second, $f_1$ belongs to shallow features, containing limited information.

We choose $f_2, f_3, f_4$ as inputs to MSFI and employ a two-stage fusion strategy for feature integration. In the first stage, we perform feature fusion from different stages using element-wise multiplication, as shown in the \autoref{eq:msfi}, where $\otimes$ represents element-wise multiplication, and $\Gamma_{\uparrow}^2$ indicates 2x upsampling. We start the computation from the feature $f_4'$ with the least computational demand, gradually passing its depth features to shallower $f_3'$ and $f_2'$. After these operations, we obtain $f_2', f_3', f_4'$. In the second stage, focusing on feature integration, we choose the concatenation operation. Similarly, starting from $f_4''$, we gradually pass it to $f_3''$ and $f_2''$.

\begin{equation}
    \left\{
    \begin{array}{l}
        f_4' = f_4 \otimes \Gamma_{\downarrow}^2(f_3)\\
        f_3' = f_3 \otimes \Gamma_{\downarrow}^2(f_2) \otimes \Gamma_{\uparrow}^2(f_4')\\
        f_2' = f_2 \otimes \Gamma_{\uparrow}^2(f_3')\\
        f_4'' = [f_4', \Gamma_{\downarrow}^2(f_3')]\\
        f_3'' = [f_3', \Gamma_{\downarrow}^2(f_2'), \Gamma_{\uparrow}^2(f_4'')]\\
        f_2'' = [f_2', \Gamma_{\uparrow}^2(f_3'')]
    \end{array}
    \right.
    \label{eq:msfi}
\end{equation}

Through the aforementioned computations, $f_2''$ has already integrated features from multiple scales. After upsampling, we input it into the decoder for mask computation. Additionally, we continue to use skip connections to concatenate the initial input and features from the next stage with $f_1, f_2'', f_3'', f_4''$, obtaining $z_1, z_2, z_3, z_4$, as shown in the \autoref{eq:z}.

\begin{equation}
    \left\{
    \begin{array}{l}
        z_4=[f_4'',f4]\\
        z_3=[f_3'',f3,\Gamma_{\uparrow}^2(z_4)]\\
        z_2=[f_2'',f2,\Gamma_{\uparrow}^2(z_3)]\\
        z_1=f_1
    \end{array}
    \right.
    \label{eq:z}
\end{equation}

\subsection{Feature extraction through autonomous kernel selection}
After going through the feature fusion and integration stage, we proceed to the feature extraction step. In this process, we designed the multi-activation selective kernel module, which selectively chooses different receptive fields to achieve feature extraction at different granularities from the features obtained in the previous step.

\subsubsection{Multi-activation convolution}
As shown in \autoref{fig:main}, we need to apply a stack of MSKM structures to feature extraction. This step involves a massive amount of computation, so we need to design a simple and efficient feature extraction operation for it. Therefore, we chose convolutional operations instead of transformers for their simplicity and efficiency, although traditional convolutional operations were still far from being concise.

\begin{figure*}[htbp]
\centering
\includegraphics[]{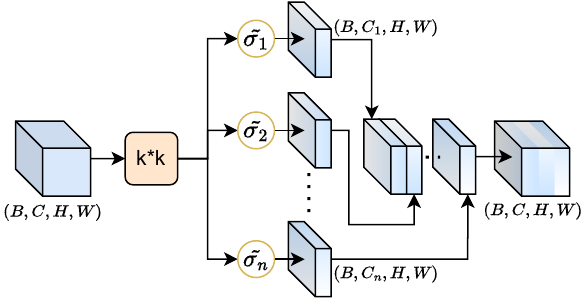}
\caption{\centering{Multi-Activation Convolution}}
\label{fig:mac}
\end{figure*}

Inspired by GhostNet \cite{ghostnet}, we observed the features in our model and identified significant redundancy. The Ghost-Module in GhostNet achieves feature expansion through linear transformations, reducing computational load. We devised a more elegant approach: Multi-Activation Convolution (MAC), as illustrated in the \autoref{fig:mac}.

We start by using a convolution operation to extract an initial set of features from the input feature $z$. Subsequently, we apply several activation functions with significant transformation differences to map the initial features. The resulting features are then concatenated along the channel dimension to obtain the feature $z'$. The calculation process is depicted in the \autoref{eq:mac1} and \autoref{eq:mac2}, where N represents the number of activation functions used, $\alpha_n$ is a trainable weight for $nth$ branch and $s_n$ signifies the features extracted by applying the nth activation function. The extracted features have the same dimensions as the input features, enabling plug-and-play functionality that can be utilized in various other models. Moreover, the operations used for initial feature extraction in the model can be flexibly chosen based on specific needs, including regular convolution, dilated convolution, or even self-attention \cite{sa}.

\begin{equation}
    s_n=\alpha_n\tilde{\sigma_n}(\mathcal{F}^{(C,\frac{C}{N},k,k)}(z)\\
    \label{eq:mac1}
\end{equation}

\begin{equation}
    z'=[s_1,s_2,\dots,s_n]\\
    \label{eq:mac2}
\end{equation}

\subsubsection{Multi-activation selective kernel module}
\begin{figure*}[htbp]
\centering
\includegraphics[width=\textwidth]{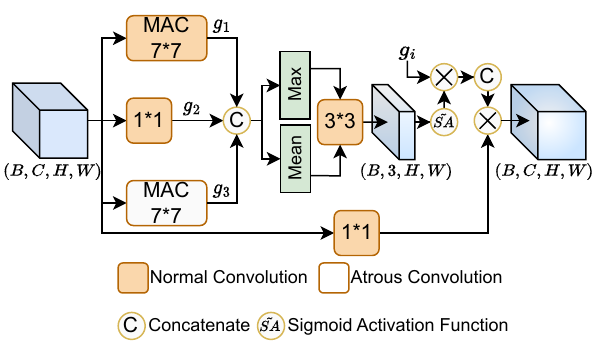}
\caption{\centering{Multi-activation Selective Kernel Module}}
\label{fig:mskm}
\end{figure*}

Camouflaged objects exhibit a unique characteristic of being generally large in size yet containing intricate details. Additionally, due to partial occlusions, camouflaged objects can be segmented into fragmented regions. Therefore, it is crucial for the model to possess multi-scale perceptual capabilities. In the process of feature extraction, the size of the receptive field plays a pivotal role. However, due to the significant variations among different camouflaged objects, a fixed receptive field size may not adequately meet the requirements of feature extraction. To address this challenge, we introduce the MSKM, as illustrated in the \autoref{fig:mskm}. This module can dynamically adapt the receptive field size based on the input image, thereby achieving a balance in extracting features at both coarse and fine scales.

In the MSKM, the input features undergo feature extraction through two MAC modules and a $1\times1$ convolutional layer, as expressed in the \autoref{eq:mskm1}. One MAC module employs a regular convolutional kernel, while the other uses a dilated convolutional kernel. The receptive field of the latter is approximately twice that of the former. As for the detail kernel size, we found that in situations with an identical receptive field, a large convolutional kernel exhibits superior edge pixel mapping compared to multiple stacked small convolutional kernels. Therefore, our model opts for the use of large convolutional kernels. Then, we concatenate these three features together.

\begin{equation}
    \left \{
        \begin{array}{l}
            g_1=\mathcal{G}_{a}^{7\times7}(z')\\
            g_2=\mathcal{F}^{1\times1}(z')\\
            g_3=\mathcal{G}_{n}^{7\times7}(z')\\
            g=[g_1,g_2,g_3]
        \end{array}
    \right .
    \label{eq:mskm1}
 \end{equation}

Subsequently, we compute the maximum and average values of the feature $g$, and transform the features into a 3-channel feature using a $3 \times 3$ convolutional layer. In the channel dimension, we apply three sigmoid activation functions for mapping, resulting in our selection $\mathcal{S}$, as shown in \autoref{eq:mskm2}. 
\begin{equation}
    \mathcal{S}=\title{SA}(\mathcal{F}^{3\times3}([max(g),mean(g)]))
    \label{eq:mskm2}
\end{equation}
We multiply $\mathcal{S}$ with the corresponding $g$ and concatenate the results. Finally, the product is obtained by multiplying with the features obtained using a $1\times1$ convolutional layer, yielding the final result. The above calculation process is shown in the \autoref{eq:mskm3} and \autoref{eq:mskm4}

\begin{equation}
    \left \{
        \begin{array}{l}
            g_1'=\mathcal{S}_1\times g_1\\
            g_2'=\mathcal{S}_2\times g_2\\
            g_3'=\mathcal{S}_3\times g_3\\
            g'=[g_1',g_2',g_3']
        \end{array}
    \right .
    \label{eq:mskm3}
\end{equation}

\begin{equation}
    m=\mathcal{F}^{1\times1}(z') \times g'
    \label{eq:mskm4}
\end{equation}

\subsection{Dual-mask strategy using spatial broadcast decoder}
After the extraction of all the necessary features, we proceed to the feature decoding stage. As depicted in \autoref{fig:main}, our model comprises a total of three decoders, consisting of two conventional decoders and one spatial broadcast decoder \cite{sbd}. The U-Net architecture is employed as the standard decoder represented as \autoref{eq:unet}, where $Conv$ represents a convolutional layer, $MaxPool$ is a max-pooling layer, $UpConv$ is an upsampling convolutional layer, $Concat$ denotes the concatenation operation, and the superscript in the upper right corner denotes the number of repetitions for that operation.

\begin{equation}
\left\{
    \begin{aligned}
        &\text{Downsampling:} \quad ((\text{Conv})^2  \rightarrow \text{MaxPool})^4 \\
        &\text{Bottleneck:} \quad (\text{Conv})^3 \\
        &\text{Upsampling:} \quad (\text{UpConv} \rightarrow \text{Concat} \rightarrow \text{Conv} )^4
    \end{aligned}
    \label{eq:unet}
\right.
\end{equation}

In the two regular decoders mentioned above, one is responsible for the final decoding of the model output, while the other is used to decode the output $f_2''$ from MSFI, generating a coarse mask denoted as $\mathcal{M}_c$. Both of these processes are directly supervised by the ground truth. Through the generation process of $\mathcal{M}_c$, we guide MSFI to retain useful features as much as possible during feature fusion.

However, having only a coarse-grained mask is far from sufficient. The real challenge lies in the generation of the fine mask. Therefore, we introduce the SBD with some modifications, adding a $1 \times 1$ convolutional layer and a $3 \times 3$ convolutional layer at its end, as shown in Algorithm \ref{algorithm}. When the encoder's final output $f_x$ is input into SBD, it is broadcasted to each pixel in the spatial domain, further incorporating spatial positional features. At this point, each pixel has global context. We use a $1 \times 1$ convolutional layer to independently calculate the feature for each pixel. Following that, we employ a $3 \times 3$ convolutional layer to fuse the features of adjacent pixels, thereby achieving the generation of the fine-grained mask $\mathcal{M}_f$. It is crucial to note that the supervision in this step involves the difference between the ground truth and $\mathcal{M}_c$.

\begin{algorithm}
\caption{Spatial Broadcast Decoder}
\textbf{Input:}
\begin{itemize}
    \item latents $z \in \mathbb{R}^k$, $w$ (width), $h$ (height)
\end{itemize}

\textbf{Output:}
\begin{itemize}
    \item tiled latents $z_{\text{sb}} \in \mathbb{R}^{h \times w \times (k+2)}$
\end{itemize}
\begin{algorithmic}[1]
    \State $zb \gets \text{tile}(z, (h, w, 1))$
    \State $x \gets \text{linspace}(-1, 1, w)$
    \State $y \gets \text{linspace}(-1, 1, h)$
    \State $xb, yb \gets \text{meshgrid}(x, y)$
    \State $z_{\text{sb}} \gets \text{concat}([zb, xb, yb], \text{axis}=-1)$
    \State $\mathcal{M}_f \gets \mathcal{F}^{3\times3}(\mathcal{F}^{1\times1}(z_{\text{sb}} ))$

\end{algorithmic}
\label{algorithm}
\end{algorithm}
\section{Experiments}
\subsection{Datasets and evaluation metrics}
When conducting the experiments, we have chosen CAMO \cite{camo}, CHAMELEON \cite{chameleon}, COD10K \cite{sinet}, and NC4K \cite{nc4k} as benchmarks to validate our methodology. These datasets are specifically curated for the task of camouflage object recognition and have been widely employed to evaluate algorithms' performance in this domain. The CAMO dataset comprises images from various scenarios where the target objects seamlessly blend into their surroundings, thereby posing a formidable challenge. CHAMELEON dataset, on the other hand, emphasizes target objects with textures and colors highly resembling their surroundings, thereby increasing recognition complexity. COD10K dataset provides additional challenging scenes involving camouflage and occlusion scenarios, aimed at assessing algorithm robustness. Finally, NC4K dataset encompasses a diverse range of natural scenes where the target objects exhibit high resemblance to the environment, thereby intensifying the recognition task. Through the selection of these datasets, our aim is to comprehensively evaluate the performance of our methodology across diverse scenarios and ascertain its efficacy and generalization capabilities in camouflage object recognition tasks.

We have selected four evaluation metrics, namely Mean Absolute Error ($\mathcal{M}$), S-measure ($S_\alpha$), Adaptive E-measure ($E_\xi$), and F-measure ($F_\beta$), to assess the performance of our camouflage object detection algorithms. $\mathcal{M}$ quantifies the average absolute error between predicted and actual values, serving as a common metric for assessing accuracy, which is the lower the better. $S_\alpha$ evaluates structural similarity, effectively gauging algorithm performance in recognizing camouflage objects. $E_\xi$ considers both pixel-level similarity and image-level statistical information, aligning with human visual perception and thus suitable for our task. $F_\beta$, a harmonic mean of precision and recall, offers a comprehensive evaluation of algorithm performance across diverse scenarios. By utilizing these evaluation metrics, we ensure a thorough and accurate assessment of our camouflage object detection algorithms.

\subsection{Implementation details}
We chose SwinV2 as the backbone for our model, with input images resized to $384\times384$ pixels. Adam optimizer was selected with a learning rate of 0.001 and a weight decay of 0.0001. For deep supervision, Dynamic Difficulty Aware Loss \cite{fsnet} was employed for the masks generated by the three decoders, while the feature mapping output by MSKI was trained using the sum of Mean Intersection over Union Loss and Binary Cross-Entropy Loss as the loss function. The batch size was set to 8, and a total of 100 epochs were trained with early stopping implemented to prevent overfitting.

\subsection{Comparison with the state-of-the-arts}
Our method was evaluated for performance on four datasets, as shown in the \autoref{tab:result}, and achieved SOTA results across all four datasets, particularly exhibiting comprehensive superiority on CAMO and COD10K.

On the CAMO dataset, compared to the high-performing FSNet, our model demonstrated overall superiority, primarily attributed to CoFiNet's ability to perceive multi-scale details, enabling it to adapt to various complex resolutions and complexities of scenes.

On the CHAMELEON dataset, our model achieved a joint first place score in the $\mathcal{M}$ metric, but lagged slightly behind JCNet in other metrics. The main reason for this is that JCNet utilizes contrastive learning methods, enhancing the model's ability to resolve similar colors and textures, enabling it to better distinguish between camouflage objects and backgrounds with highly similar colors and textures.

The COD10K dataset provides diverse scenes and challenges, where target objects may be occluded or camouflaged by other objects. Our model achieved the best results on all metrics. CoFiNet's dual-mask strategy effectively strengthens the model's ability to detect details, leading to performance improvements.

On the NC4K dataset, CoFiNet narrowly ranked second in the $F_\beta$ metric but ranked first in other metrics, reflecting CoFiNet's performance and generalization ability in real-world scenarios.

To visually illustrate the comparison of our experimental results, we provide a Visual comparison with previous SOTA in the \autoref{secA1}.

\begin{table}[htbp]
  \centering
  \resizebox{\textwidth}{!}{
    \begin{tabular}{ccccc|cccc|cccc|cccc}  
      \hline  
      \multirow{2}*{Method} & \multicolumn{4}{c}{CAMO \cite{camo}}  & \multicolumn{4}{c}{CHAMELEON \cite{chameleon}}  & \multicolumn{4}{c}{COD10K \cite{sinet}} & \multicolumn{4}{c}{NC4K \cite{nc4k}} \\
      \cmidrule{2-17}
      & $\mathcal{S}_{\alpha}\uparrow$ & $F_{\beta} \uparrow$ & $E_{\xi}\uparrow$ & $\mathcal{M}\downarrow$ & $\mathcal{S}_{\alpha}\uparrow$ & $F_{\beta} \uparrow$ & $E_{\xi}\uparrow$ & $\mathcal{M}\downarrow$ & $\mathcal{S}_{\alpha}\uparrow$ & $F_{\beta} \uparrow$ & $E_{\xi}\uparrow$ & $\mathcal{M}\downarrow$ & $\mathcal{S}_{\alpha}\uparrow$ & $F_{\beta} \uparrow$ & $E_{\xi}\uparrow$ & $\mathcal{M}\downarrow$   \\  
      \hline  
      SCRN \cite{scrn}&0.779&0.705&0.796&0.090&0.876&0.787&0.889&0.042&0.789&0.651&0.817&0.047&0.832&0.759&0.855&0.059\\
      CSNet\cite{csnet}&0.771&0.705&0.795&0.092&0.876&0.787&0.889&0.042&0.778&0.635&0.810&0.047&0.819&0.748&0.845&0.061\\
      EGNet\cite{egnet}&0.735&0.650&0.753&0.102&0.856&0.763&0.877&0.049&0.748&0.587&0.776&0.053&0.796&0.718&0.830&0.067\\
      LSR\cite{nc4k}&0.793&0.725&0.826&0.085&0.893&0.839&0.938&0.033&0.793&0.685&0.868&0.041&0.839&0.779&0.883&0.053\\
      UJSC\cite{ujsc}&0.803&0.759&0.853&0.076&0.894&0.848&0.943&0.030&0.817&0.726&0.892&0.035&0.842&0.806&0.898&0.047\\
      MGL\cite{mgl}&0.775&0.726&0.812&0.088&0.893&0.834&0.918&0.030&0.814&0.711&0.852&0.035&0.833&0.782&0.867&0.052\\
      PFNet\cite{pfnet}&0.782&0.744&0.840&0.085&0.882&0.826&0.922&0.033&0.800&0.700&0.875&0.040&0.829&0.782&0.886&0.053\\
      PraNet\cite{pranet}&0.769&0.711&0.825&0.094&0.860&0.790&0.908&0.044&0.790&0.672&0.861&0.045&0.822&0.763&0.877&0.059\\
      SINet\cite{sinet}&0.745&0.702&0.804&0.092&0.872&0.827&0.936&0.034&0.776&0.679&0.864&0.043&0.810&0.772&0.873&0.057\\
      SINet-V2\cite{sinetv2}&0.820&0.782&0.882&0.070&0.888&0.835&0.942&0.030&0.815&0.718&0.887&0.037&0.847&0.805&0.903&0.048\\
      JCNet\cite{jcnet}&0.850&0.827&0.913&0.054&\textbf{0.906}&\textbf{0.872}&[0.956]&\textbf{0.021}&0.852&0.788&0.927&0.026&0.876&0.850&0.931&0.035\\
      FSNet\cite{fsnet}&[0.880]&[0.861]&[0.933]&[0.041]&[0.905]&0.868&\textbf{0.963}&[0.022]&[0.870]&[0.810]&[0.938]&[0.023]&[0.891]&\textbf{0.866}&[0.940]&[0.031]\\
      \hline  
      CoFiNet&\textbf{0.887}&\textbf{0.877}&\textbf{0.941}&\textbf{0.038} &[0.905]&[0.870]&0.953&\textbf{0.021}&\textbf{0.873}&\textbf{0.814}&\textbf{0.943}&\textbf{0.021}&\textbf{0.894}&[0.865]&\textbf{0.944}&\textbf{0.029}\\
      \hline  

    \end{tabular}
  }
  \caption{\centering{Experimental result on multiple dataset}}
  \label{tab:result}
\end{table}

\subsection{Ablation study}
To demonstrate the effectiveness of our proposed method, we conducted ablation experiments on the COD10K dataset. The ablation comparison mainly consists of three parts: the feature fusion module, the feature extraction module, and the finer decoder. In the feature fusion module, we compared our MSKM with the cross-connection decoder (CCD) of FSNet. For the feature fusion module, we compared our MSKM with DenseASPP \cite{denseaspp}. Regarding the finer decoder, we compared the experimental results with and without the addition of the spatial broadcast decoder. The results of the ablation experiments are shown in \autoref{tab:ablation}.

From the table, we observe that our final model achieves the most effective performance across all metrics, validating the effectiveness of each module in our model. Comparing No. 5 with No. 6 and No. 1 with No. 2, we observe a significant improvement in the $\mathcal{M}$ metric when SBD is added, demonstrating that the adoption of SBD's dual-mask strategy effectively enhances the model's detection and segmentation capabilities. No. 2 versus No. 6 and No. 4 versus No. 6 respectively reflect the effectiveness of the MSKM and MSFI modules.

\begin{table}[htbp]
  \centering
  \resizebox{\textwidth}{!}{
    \begin{tabular}{cccccccccc}  
      \hline  
      \multirow{2}*{No.} & \multicolumn{2}{@{}c@{}}{Fusion module}  & \multicolumn{2}{@{}c@{}}{Extraction module}  & Finer decoder & \multicolumn{4}{c}{COD10K} \\
        \cmidrule{2-3}\cmidrule{4-5}\cmidrule{6-6}\cmidrule{7-10}%
      & MSFI& CCD & MSKM&DenseASPP&SBD & $\mathcal{S}_{\alpha}\uparrow$ & $F_{\beta} \uparrow$ & $E_{\xi}\uparrow$ & $\mathcal{M}\downarrow$   \\  
      \hline  
1&\checkmark&&&\checkmark&&0.859&0.801&0.929&0.025\\
2&\checkmark&&&\checkmark&\checkmark&0.865&0.812&0.936&0.022\\
3&&\checkmark&\checkmark&&&0.862&0.805&0.933&0.023\\
4&&\checkmark&\checkmark&&\checkmark&0.870&0.807&0.938&0.022\\
5&\checkmark&&\checkmark&&&0.869&0.808&0.934&0.023\\
6[CoFiNet]&\checkmark&&\checkmark&&\checkmark&\textbf{0.873}&\textbf{0.814}&\textbf{0.943}&\textbf{0.021}\\
      \hline  

    \end{tabular}
  }
  \caption{\centering{Ablation study}}
  \label{tab:ablation}
\end{table}

\subsection{Case study}
\begin{figure}[htbp]
    \centering
    \includegraphics[width=0.9\textwidth]{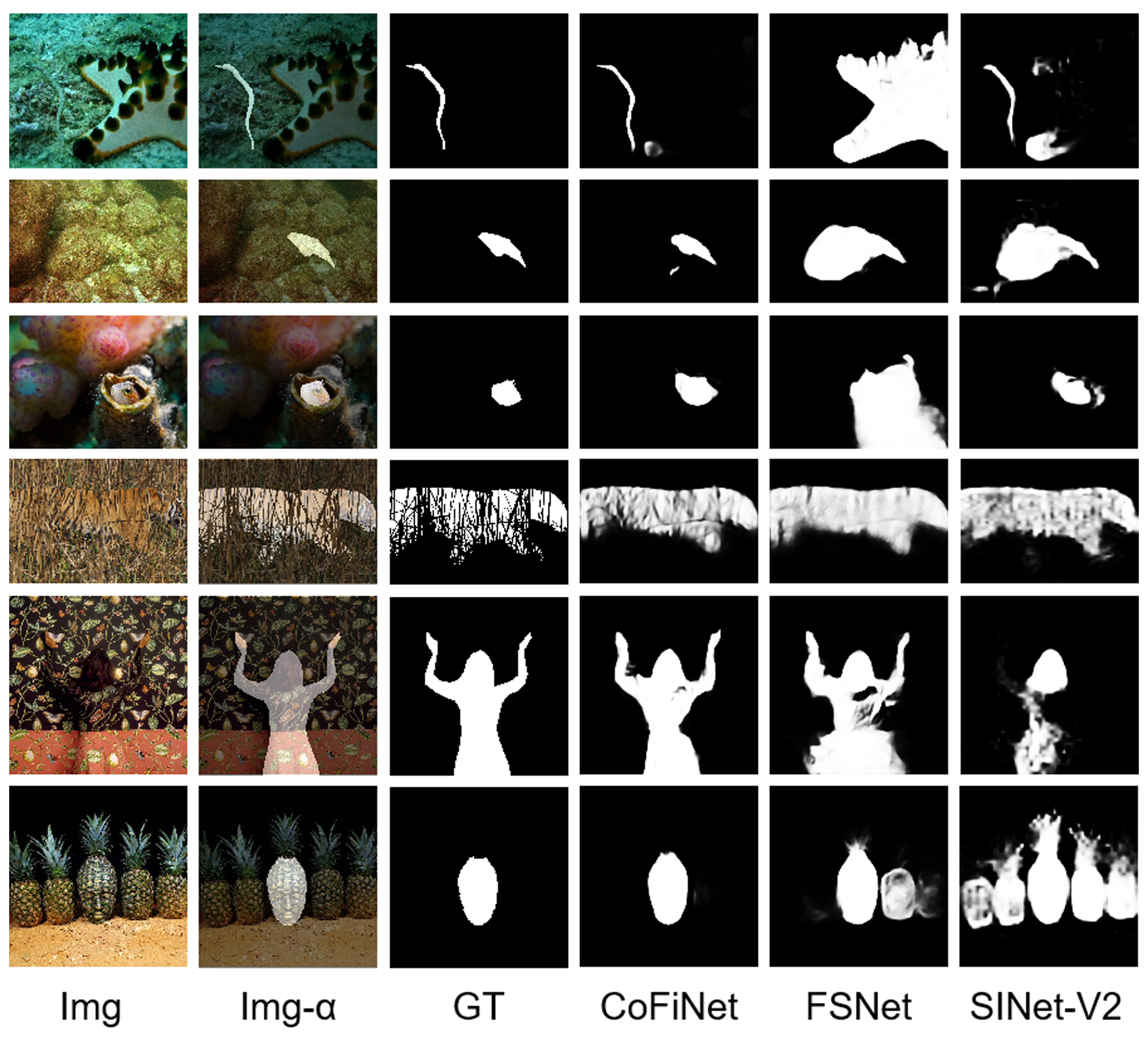}
    \caption{\centering{Visual comparisons.}}
    \label{fig:casestudy}
\end{figure}

To further demonstrate the performance of CoFiNet, we conducted a case study on selected samples, as shown in the \autoref{fig:casestudy}. The images are divided into 6 rows and 6 columns, with each row representing a case. The first column displays the original images, the second column shows the overlay of the original image and the ground truth, the third column presents the ground truth, and the fourth to sixth columns display the detection results of three methods.

In the first example, CoFiNet successfully segmented the worm and the starfish, while SINetV2's segmentation result was mediocre, and FSNet directly treated the salient object as the segmentation result. This example demonstrates our model's capability to distinguish between salient and camouflaged objects effectively. In the second case, where the camouflaged object and the background exhibit high texture and color similarity, only CoFiNet successfully detected it, showcasing the model's perception of texture and color. The third example illustrates the detection of small-sized camouflaged objects, where CoFiNet achieved good segmentation results due to its excellent multi-scale perception capability. In the fourth scenario, where the camouflaged object is occluded, particularly by sparse vegetation, CoFiNet achieved finer segmentation of grass details compared to other methods, highlighting the effectiveness of our coarse-fine strategy. The fifth and sixth examples depict body painting and digital synthesis images, where our model also performed excellently, demonstrating its generalization capability and robustness.

The case study above verifies CoFiNet's ability to handle complex camouflaged objects across various scenes and scales, showcasing the effectiveness and reliability of the model.

\section{Conclusion}
In this paper, we propose CoFiNet, which further improves the detection of camouflaged objects from a multi-scale perspective, progressing from coarse to fine. For feature fusion and integration, we introduce the MSFI module. Our proposed MSKM effectively enhances the model's feature extraction capability by utilizing variable-sized receptive fields. The dual-mask strategy we propose further enhances the model's ability to reconstruct detailed features. We conducted experiments on four datasets, and our model achieved state-of-the-art performance on all of them. Additionally, we performed ablation experiments on the COD10K dataset to validate the effectiveness of each module we proposed. The extensive case studies further demonstrate the model's generalization capability and robustness.

\section{Competing Interests}
The authors have no competing interests to declare that are relevant to the content of this article.
\section{Authors contribution statement}
Cunhan Guo provided the overall idea of the article and completed the experiments and paper writing. Heyan Huang provided the necessary computing power for the experiment and also guided the writing and revision of the paper.
\section{Ethical and informed consent for data used}
The data used in this paper are from the public data set, which has been quoted in the paper. And there are no ethical issues with these data.
\section{Data Availability Statement}
The data that support the findings of this study are available from the authors, upon reasonable request.

\begin{appendices}
\section{Visual comparison}\label{secA1}
\begin{figure}[htbp]
    \centering
    \includegraphics[width=0.65\textwidth]{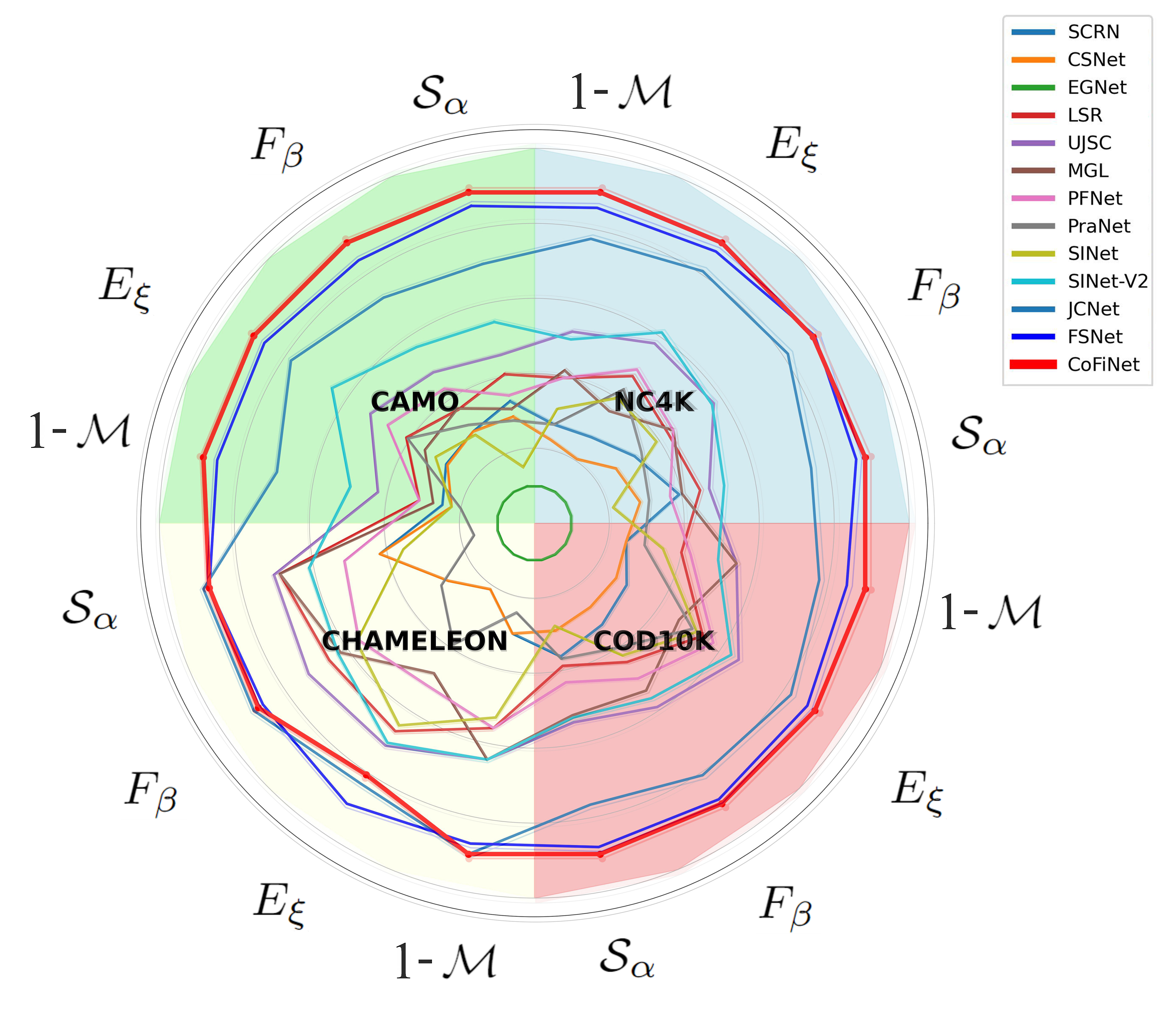}
    \caption{\centering{Compare with other method.}}
    \label{fig:comparison}
\end{figure}
As shown in the \autoref{fig:comparison}, we visually displayed the evaluation metrics on the four datasets, with each metric independently normalized for averaging. The four different background colors represent the four datasets, while the 13 lines represent 13 different methods. Results closer to the outer circle indicate better performance. Our results, represented by the red line in the image, encircle other lines in multiple directions, demonstrating the outstanding performance of CoFiNet.

\section{Failure cases}\label{secB2}
\begin{figure}[htbp]
    \centering
    \includegraphics[width=0.9\textwidth]{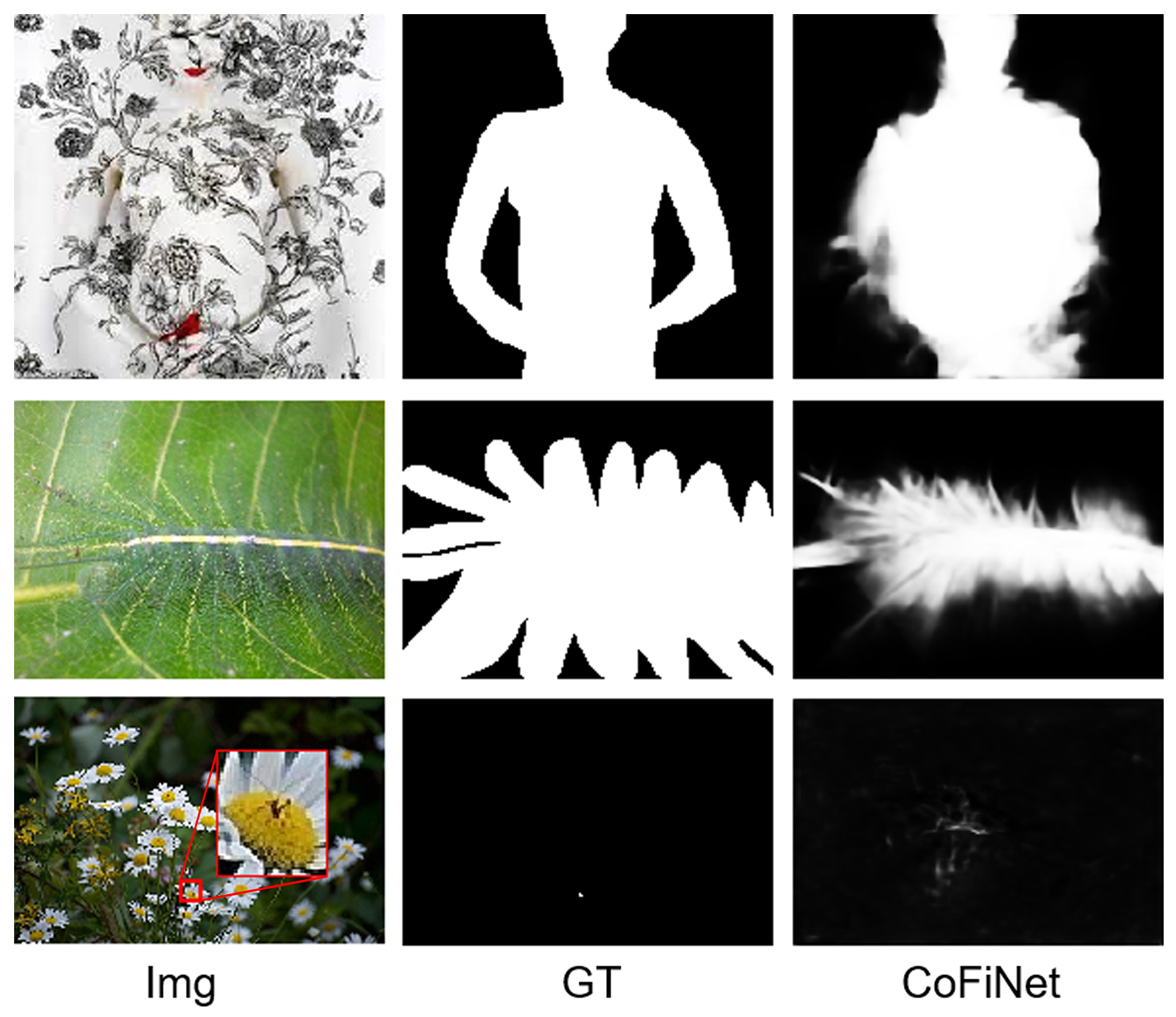}
    \caption{\centering{Failure cases.}}
    \label{fig:worst}
\end{figure}

The \autoref{fig:worst} showcase some failure cases of CoFiNet. 

In the first case, a body painting with intricate details over a pure white base color is depicted. While our model can roughly detect the object, the segmentation lacks fine granularity. The main reason for this is the high similarity in color, which limits the model's discriminative ability. Further optimization of the model's color perception capability could address such issues in the future. 

In the second case, the ground truth labels consider all the holes within the fine edges as part of the camouflaged object. However, due to the limited number of relevant samples, our model identifies only the solid parts as camouflaged objects, resulting in suboptimal segmentation. 

The third case involves the detection of a tiny insect within a cluster of flowers. Since our model's input is limited to $384 \times 384$ dimensions, details at this scale may not be detectable when the original image is resized to fit this dimension. To address this, automatic local cropping before detection could be applied for specific scenes in the future.

\end{appendices}

\bibliography{sn-bibliography}


\begin{thebibliography}{49}
\ifx \bisbn   \undefined \def \bisbn  #1{ISBN #1}\fi
\ifx \binits  \undefined \def \binits#1{#1}\fi
\ifx \bauthor  \undefined \def \bauthor#1{#1}\fi
\ifx \batitle  \undefined \def \batitle#1{#1}\fi
\ifx \bjtitle  \undefined \def \bjtitle#1{#1}\fi
\ifx \bvolume  \undefined \def \bvolume#1{\textbf{#1}}\fi
\ifx \byear  \undefined \def \byear#1{#1}\fi
\ifx \bissue  \undefined \def \bissue#1{#1}\fi
\ifx \bfpage  \undefined \def \bfpage#1{#1}\fi
\ifx \blpage  \undefined \def \blpage #1{#1}\fi
\ifx \burl  \undefined \def \burl#1{\textsf{#1}}\fi
\ifx \doiurl  \undefined \def \doiurl#1{\url{https://doi.org/#1}}\fi
\ifx \betal  \undefined \def \betal{\textit{et al.}}\fi
\ifx \binstitute  \undefined \def \binstitute#1{#1}\fi
\ifx \binstitutionaled  \undefined \def \binstitutionaled#1{#1}\fi
\ifx \bctitle  \undefined \def \bctitle#1{#1}\fi
\ifx \beditor  \undefined \def \beditor#1{#1}\fi
\ifx \bpublisher  \undefined \def \bpublisher#1{#1}\fi
\ifx \bbtitle  \undefined \def \bbtitle#1{#1}\fi
\ifx \bedition  \undefined \def \bedition#1{#1}\fi
\ifx \bseriesno  \undefined \def \bseriesno#1{#1}\fi
\ifx \blocation  \undefined \def \blocation#1{#1}\fi
\ifx \bsertitle  \undefined \def \bsertitle#1{#1}\fi
\ifx \bsnm \undefined \def \bsnm#1{#1}\fi
\ifx \bsuffix \undefined \def \bsuffix#1{#1}\fi
\ifx \bparticle \undefined \def \bparticle#1{#1}\fi
\ifx \barticle \undefined \def \barticle#1{#1}\fi
\bibcommenthead
\ifx \bconfdate \undefined \def \bconfdate #1{#1}\fi
\ifx \botherref \undefined \def \botherref #1{#1}\fi
\ifx \url \undefined \def \url#1{\textsf{#1}}\fi
\ifx \bchapter \undefined \def \bchapter#1{#1}\fi
\ifx \bbook \undefined \def \bbook#1{#1}\fi
\ifx \bcomment \undefined \def \bcomment#1{#1}\fi
\ifx \oauthor \undefined \def \oauthor#1{#1}\fi
\ifx \citeauthoryear \undefined \def \citeauthoryear#1{#1}\fi
\ifx \endbibitem  \undefined \def \endbibitem {}\fi
\ifx \bconflocation  \undefined \def \bconflocation#1{#1}\fi
\ifx \arxivurl  \undefined \def \arxivurl#1{\textsf{#1}}\fi
\csname PreBibitemsHook\endcsname

\bibitem[\protect\citeauthoryear{Singh et~al.}{2013}]{early-1}
\begin{barticle}
\bauthor{\bsnm{Singh}, \binits{S.K.}},
\bauthor{\bsnm{Dhawale}, \binits{C.A.}},
\bauthor{\bsnm{Misra}, \binits{S.}}:
\batitle{Survey of object detection methods in camouflaged image}.
\bjtitle{IERI Procedia}
\bvolume{4},
\bfpage{351}--\blpage{357}
(\byear{2013})
\doiurl{10.1016/j.ieri.2013.11.050} .
\bcomment{2013 International Conference on Electronic Engineering and Computer Science (EECS 2013)}
\end{barticle}
\endbibitem

\bibitem[\protect\citeauthoryear{{Wendi Hou Jinping Li}}{2011}]{early-2}
\begin{barticle}
\bauthor{\bsnm{{Wendi Hou Jinping Li}}, \binits{J.Y.Y.H.}}:
\batitle{Detection of the mobile object with camouflage color under dynamic background based on optical flow}.
\bjtitle{Procedia Engineering}
\bvolume{15},
\bfpage{2201}--\blpage{2205}
(\byear{2011})
\doiurl{10.1016/j.proeng.2011.08.412} .
\bcomment{CEIS 2011}
\end{barticle}
\endbibitem

\bibitem[\protect\citeauthoryear{Zhang et~al.}{2017}]{early-3}
\begin{barticle}
\bauthor{\bsnm{Zhang}, \binits{X.}},
\bauthor{\bsnm{Zhu}, \binits{C.}},
\bauthor{\bsnm{Wang}, \binits{S.}},
\bauthor{\bsnm{Liu}, \binits{Y.}},
\bauthor{\bsnm{Ye}, \binits{M.}}:
\batitle{A bayesian approach to camouflaged moving object detection}.
\bjtitle{{IEEE} Trans. Circuits Syst. Video Technol.}
\bvolume{27}(\bissue{9}),
\bfpage{2001}--\blpage{2013}
(\byear{2017})
\doiurl{10.1109/TCSVT.2016.2555719}
\end{barticle}
\endbibitem

\bibitem[\protect\citeauthoryear{Zhang and Zhu}{2015}]{early-4}
\begin{bchapter}
\bauthor{\bsnm{Zhang}, \binits{X.}},
\bauthor{\bsnm{Zhu}, \binits{C.}}:
\bctitle{Camouflage modeling for moving object detection}.
In: \bbtitle{{IEEE} China Summit and International Conference on Signal and Information Processing, ChinaSIP 2015, Chengdu, China, July 12-15, 2015},
pp. \bfpage{249}--\blpage{253}.
\bpublisher{{IEEE}},
\blocation{Chengdu, China}
(\byear{2015}).
\doiurl{10.1109/CHINASIP.2015.7230401} .
\burl{https://doi.org/10.1109/ChinaSIP.2015.7230401}
\end{bchapter}
\endbibitem

\bibitem[\protect\citeauthoryear{Cheng et~al.}{2018}]{infrared}
\begin{barticle}
\bauthor{\bsnm{Cheng}, \binits{X.-p.}},
\bauthor{\bsnm{Zhao}, \binits{D.-p.}},
\bauthor{\bsnm{Yu}, \binits{Z.-j.}},
\bauthor{\bsnm{Zhang}, \binits{J.-h.}},
\bauthor{\bsnm{Bian}, \binits{J.-t.}},
\bauthor{\bsnm{Yu}, \binits{D.-b.}}:
\batitle{Effectiveness evaluation of infrared camouflage using image saliency}.
\bjtitle{Infrared Physics \&amp; Technology}
\bvolume{95},
\bfpage{213}--\blpage{221}
(\byear{2018})
\doiurl{10.1016/j.infrared.2018.11.001}
\end{barticle}
\endbibitem

\bibitem[\protect\citeauthoryear{Kumar and Ghosh}{2016}]{hyperspectral}
\begin{barticle}
\bauthor{\bsnm{Kumar}, \binits{V.}},
\bauthor{\bsnm{Ghosh}, \binits{J.K.}}:
\batitle{Camouflage detection using mwir hyperspectral images}.
\bjtitle{Journal of the Indian Society of Remote Sensing}
\bvolume{45}(\bissue{1}),
\bfpage{139}--\blpage{145}
(\byear{2016})
\doiurl{10.1007/s12524-016-0555-8}
\end{barticle}
\endbibitem

\bibitem[\protect\citeauthoryear{Wang et~al.}{2024}]{polarization}
\begin{barticle}
\bauthor{\bsnm{Wang}, \binits{X.}},
\bauthor{\bsnm{Ding}, \binits{J.}},
\bauthor{\bsnm{Zhang}, \binits{Z.}},
\bauthor{\bsnm{Xu}, \binits{J.}},
\bauthor{\bsnm{Gao}, \binits{J.}}:
\batitle{Ipnet: Polarization-based camouflaged object detection via dual-flow network}.
\bjtitle{Eng. Appl. Artif. Intell.}
\bvolume{127}(\bissue{Part {A}}),
\bfpage{107303}
(\byear{2024})
\doiurl{10.1016/J.ENGAPPAI.2023.107303}
\end{barticle}
\endbibitem

\bibitem[\protect\citeauthoryear{Liang et~al.}{2024}]{survey}
\begin{barticle}
\bauthor{\bsnm{Liang}, \binits{Y.}},
\bauthor{\bsnm{Qin}, \binits{G.}},
\bauthor{\bsnm{Sun}, \binits{M.}},
\bauthor{\bsnm{Wang}, \binits{X.}},
\bauthor{\bsnm{Yan}, \binits{J.}},
\bauthor{\bsnm{Zhang}, \binits{Z.}}:
\batitle{A systematic review of image-level camouflaged object detection with deep learning}.
\bjtitle{Neurocomputing}
\bvolume{566},
\bfpage{127050}
(\byear{2024})
\doiurl{10.1016/J.NEUCOM.2023.127050}
\end{barticle}
\endbibitem

\bibitem[\protect\citeauthoryear{Gao and Fan}{2006}]{sod}
\begin{bchapter}
\bauthor{\bsnm{Gao}, \binits{Y.}},
\bauthor{\bsnm{Fan}, \binits{J.}}:
\bctitle{Automatic function selection for large scale salient object detection}.
In: \bbtitle{Proceedings of the 14th ACM International Conference on Multimedia},
pp. \bfpage{97}--\blpage{100}
(\byear{2006})
\end{bchapter}
\endbibitem

\bibitem[\protect\citeauthoryear{Fan et~al.}{2020}]{sinet}
\begin{bchapter}
\bauthor{\bsnm{Fan}, \binits{D.-P.}},
\bauthor{\bsnm{Ji}, \binits{G.-P.}},
\bauthor{\bsnm{Sun}, \binits{G.}},
\bauthor{\bsnm{Cheng}, \binits{M.-M.}},
\bauthor{\bsnm{Shen}, \binits{J.}},
\bauthor{\bsnm{Shao}, \binits{L.}}:
\bctitle{Camouflaged object detection}.
In: \bbtitle{2020 IEEE/CVF Conference on Computer Vision and Pattern Recognition (CVPR)},
pp. \bfpage{2774}--\blpage{2784}
(\byear{2020}).
\doiurl{10.1109/CVPR42600.2020.00285}
\end{bchapter}
\endbibitem

\bibitem[\protect\citeauthoryear{Fan et~al.}{2022}]{sinetv2}
\begin{barticle}
\bauthor{\bsnm{Fan}, \binits{D.-P.}},
\bauthor{\bsnm{Ji}, \binits{G.-P.}},
\bauthor{\bsnm{Cheng}, \binits{M.-M.}},
\bauthor{\bsnm{Shao}, \binits{L.}}:
\batitle{Concealed object detection}.
\bjtitle{IEEE Transactions on Pattern Analysis and Machine Intelligence}
\bvolume{44}(\bissue{10}),
\bfpage{6024}--\blpage{6042}
(\byear{2022})
\doiurl{10.1109/TPAMI.2021.3085766}
\end{barticle}
\endbibitem

\bibitem[\protect\citeauthoryear{Viola and Jones}{}]{viola}
\begin{botherref}
\oauthor{\bsnm{Viola}, \binits{P.}},
\oauthor{\bsnm{Jones}, \binits{M.}}:
Rapid object detection using a boosted cascade of simple features.
In: Proceedings of the 2001 IEEE Computer Society Conference on Computer Vision and Pattern Recognition. CVPR 2001.
CVPR-01.
IEEE Comput. Soc.
\doiurl{10.1109/cvpr.2001.990517} .
\url{http://dx.doi.org/10.1109/CVPR.2001.990517}
\end{botherref}
\endbibitem

\bibitem[\protect\citeauthoryear{Dalal and Triggs}{}]{hog}
\begin{botherref}
\oauthor{\bsnm{Dalal}, \binits{N.}},
\oauthor{\bsnm{Triggs}, \binits{B.}}:
Histograms of oriented gradients for human detection.
In: 2005 IEEE Computer Society Conference on Computer Vision and Pattern Recognition (CVPR’05).
IEEE.
\doiurl{10.1109/cvpr.2005.177} .
\url{http://dx.doi.org/10.1109/CVPR.2005.177}
\end{botherref}
\endbibitem

\bibitem[\protect\citeauthoryear{Jung et~al.}{2014}]{dpm}
\begin{bchapter}
\bauthor{\bsnm{Jung}, \binits{H.}},
\bauthor{\bsnm{Ju}, \binits{J.}},
\bauthor{\bsnm{Kim}, \binits{J.}}:
\bctitle{Rigid motion segmentation using randomized voting}.
In: \bbtitle{Proceedings of the IEEE Conference on Computer Vision and Pattern Recognition},
pp. \bfpage{1210}--\blpage{1217}
(\byear{2014})
\end{bchapter}
\endbibitem

\bibitem[\protect\citeauthoryear{Girshick et~al.}{2014}]{rcnn}
\begin{bchapter}
\bauthor{\bsnm{Girshick}, \binits{R.}},
\bauthor{\bsnm{Donahue}, \binits{J.}},
\bauthor{\bsnm{Darrell}, \binits{T.}},
\bauthor{\bsnm{Malik}, \binits{J.}}:
\bctitle{Rich feature hierarchies for accurate object detection and semantic segmentation}.
In: \bbtitle{Proceedings of the IEEE Conference on Computer Vision and Pattern Recognition},
pp. \bfpage{580}--\blpage{587}
(\byear{2014})
\end{bchapter}
\endbibitem

\bibitem[\protect\citeauthoryear{Lin et~al.}{2017}]{fpn}
\begin{bchapter}
\bauthor{\bsnm{Lin}, \binits{T.-Y.}},
\bauthor{\bsnm{Dollár}, \binits{P.}},
\bauthor{\bsnm{Girshick}, \binits{R.}},
\bauthor{\bsnm{He}, \binits{K.}},
\bauthor{\bsnm{Hariharan}, \binits{B.}},
\bauthor{\bsnm{Belongie}, \binits{S.}}:
\bctitle{Feature pyramid networks for object detection}.
In: \bbtitle{2017 IEEE Conference on Computer Vision and Pattern Recognition (CVPR)},
pp. \bfpage{936}--\blpage{944}
(\byear{2017}).
\doiurl{10.1109/CVPR.2017.106}
\end{bchapter}
\endbibitem

\bibitem[\protect\citeauthoryear{Redmon et~al.}{2016}]{yolo}
\begin{bchapter}
\bauthor{\bsnm{Redmon}, \binits{J.}},
\bauthor{\bsnm{Divvala}, \binits{S.}},
\bauthor{\bsnm{Girshick}, \binits{R.}},
\bauthor{\bsnm{Farhadi}, \binits{A.}}:
\bctitle{You only look once: Unified, real-time object detection}.
In: \bbtitle{Proceedings of the IEEE Conference on Computer Vision and Pattern Recognition},
pp. \bfpage{779}--\blpage{788}
(\byear{2016})
\end{bchapter}
\endbibitem

\bibitem[\protect\citeauthoryear{Liu et~al.}{2016}]{ssd}
\begin{bchapter}
\bauthor{\bsnm{Liu}, \binits{W.}},
\bauthor{\bsnm{Anguelov}, \binits{D.}},
\bauthor{\bsnm{Erhan}, \binits{D.}},
\bauthor{\bsnm{Szegedy}, \binits{C.}},
\bauthor{\bsnm{Reed}, \binits{S.}},
\bauthor{\bsnm{Fu}, \binits{C.-Y.}},
\bauthor{\bsnm{Berg}, \binits{A.C.}}:
\bctitle{Ssd: Single shot multibox detector}.
In: \bbtitle{Computer Vision--ECCV 2016: 14th European Conference, Amsterdam, The Netherlands, October 11--14, 2016, Proceedings, Part I 14},
pp. \bfpage{21}--\blpage{37}
(\byear{2016}).
\bcomment{Springer}
\end{bchapter}
\endbibitem

\bibitem[\protect\citeauthoryear{Lin et~al.}{2020}]{retinaNet}
\begin{barticle}
\bauthor{\bsnm{Lin}, \binits{T.-Y.}},
\bauthor{\bsnm{Goyal}, \binits{P.}},
\bauthor{\bsnm{Girshick}, \binits{R.}},
\bauthor{\bsnm{He}, \binits{K.}},
\bauthor{\bsnm{Dollar}, \binits{P.}}:
\batitle{Focal loss for dense object detection}.
\bjtitle{IEEE Transactions on Pattern Analysis and Machine Intelligence}
\bvolume{42}(\bissue{2}),
\bfpage{318}--\blpage{327}
(\byear{2020})
\doiurl{10.1109/tpami.2018.2858826}
\end{barticle}
\endbibitem

\bibitem[\protect\citeauthoryear{Borji et~al.}{2015}]{sod_baseline}
\begin{barticle}
\bauthor{\bsnm{Borji}, \binits{A.}},
\bauthor{\bsnm{Cheng}, \binits{M.}},
\bauthor{\bsnm{Jiang}, \binits{H.}},
\bauthor{\bsnm{Li}, \binits{J.}}:
\batitle{Salient object detection: {A} benchmark}.
\bjtitle{{IEEE} Trans. Image Process.}
\bvolume{24}(\bissue{12}),
\bfpage{5706}--\blpage{5722}
(\byear{2015})
\doiurl{10.1109/TIP.2015.2487833}
\end{barticle}
\endbibitem

\bibitem[\protect\citeauthoryear{Wang}{2022}]{usl_sod}
\begin{barticle}
\bauthor{\bsnm{Wang}, \binits{S.}}:
\batitle{Learning nonlinear feature mapping via constrained non-convex optimization for unsupervised salient object detection}.
\bjtitle{{IEEE} Access}
\bvolume{10},
\bfpage{40743}--\blpage{40752}
(\byear{2022})
\doiurl{10.1109/ACCESS.2022.3166986}
\end{barticle}
\endbibitem

\bibitem[\protect\citeauthoryear{Lu et~al.}{2023}]{seg_sod}
\begin{barticle}
\bauthor{\bsnm{Lu}, \binits{Z.}},
\bauthor{\bsnm{Liang}, \binits{H.}},
\bauthor{\bsnm{Xu}, \binits{B.}},
\bauthor{\bsnm{Liang}, \binits{R.}}:
\batitle{A progressive segmentation with weight contrast label enhancement for weakly supervised video salient object detection}.
\bjtitle{{IET} Image Process.}
\bvolume{17}(\bissue{10}),
\bfpage{2925}--\blpage{2936}
(\byear{2023})
\doiurl{10.1049/IPR2.12840}
\end{barticle}
\endbibitem

\bibitem[\protect\citeauthoryear{Huxohl}{2023}]{detect_sod}
\begin{botherref}
\oauthor{\bsnm{Huxohl}, \binits{T.}}:
Deep learning based salient object detection for the detection of stains and holes on patterned laundry.
PhD thesis,
Bielefeld University, Germany
(2023).
\url{https://pub.uni-bielefeld.de/record/2979082}
\end{botherref}
\endbibitem

\bibitem[\protect\citeauthoryear{He et~al.}{2015}]{supercnn}
\begin{barticle}
\bauthor{\bsnm{He}, \binits{S.}},
\bauthor{\bsnm{Lau}, \binits{R.W.H.}},
\bauthor{\bsnm{Liu}, \binits{W.}},
\bauthor{\bsnm{Huang}, \binits{Z.}},
\bauthor{\bsnm{Yang}, \binits{Q.}}:
\batitle{Supercnn: {A} superpixelwise convolutional neural network for salient object detection}.
\bjtitle{Int. J. Comput. Vis.}
\bvolume{115}(\bissue{3}),
\bfpage{330}--\blpage{344}
(\byear{2015})
\doiurl{10.1007/S11263-015-0822-0}
\end{barticle}
\endbibitem

\bibitem[\protect\citeauthoryear{Xing et~al.}{2023}]{multitask}
\begin{botherref}
\oauthor{\bsnm{Xing}, \binits{Y.}},
\oauthor{\bsnm{Kong}, \binits{D.}},
\oauthor{\bsnm{Zhang}, \binits{S.}},
\oauthor{\bsnm{Chen}, \binits{G.}},
\oauthor{\bsnm{Ran}, \binits{L.}},
\oauthor{\bsnm{Wang}, \binits{P.}},
\oauthor{\bsnm{Zhang}, \binits{Y.}}:
Pre-train, adapt and detect: Multi-task adapter tuning for camouflaged object detection.
CoRR
\textbf{abs/2307.10685}
(2023)
\doiurl{10.48550/ARXIV.2307.10685}
{\href{https://arxiv.org/abs/2307.10685}{{2307.10685}}}
\end{botherref}
\endbibitem

\bibitem[\protect\citeauthoryear{Lv et~al.}{2021}]{nc4k}
\begin{bchapter}
\bauthor{\bsnm{Lv}, \binits{Y.}},
\bauthor{\bsnm{Zhang}, \binits{J.}},
\bauthor{\bsnm{Dai}, \binits{Y.}},
\bauthor{\bsnm{Li}, \binits{A.}},
\bauthor{\bsnm{Liu}, \binits{B.}},
\bauthor{\bsnm{Barnes}, \binits{N.}},
\bauthor{\bsnm{Fan}, \binits{D.-P.}}:
\bctitle{Simultaneously localize, segment and rank the camouflaged objects}.
In: \bbtitle{Proceedings of the IEEE/CVF Conference on Computer Vision and Pattern Recognition},
pp. \bfpage{11591}--\blpage{11601}
(\byear{2021})
\end{bchapter}
\endbibitem

\bibitem[\protect\citeauthoryear{Le et~al.}{2021}]{multitask2}
\begin{botherref}
\oauthor{\bsnm{Le}, \binits{T.}},
\oauthor{\bsnm{Nguyen}, \binits{T.V.}},
\oauthor{\bsnm{Nie}, \binits{Z.}},
\oauthor{\bsnm{Tran}, \binits{M.}},
\oauthor{\bsnm{Sugimoto}, \binits{A.}}:
Anabranch network for camouflaged object segmentation.
CoRR
\textbf{abs/2105.09451}
(2021)
{\href{https://arxiv.org/abs/2105.09451}{{2105.09451}}}
\end{botherref}
\endbibitem

\bibitem[\protect\citeauthoryear{Ren et~al.}{2023}]{multitask3}
\begin{barticle}
\bauthor{\bsnm{Ren}, \binits{J.}},
\bauthor{\bsnm{Hu}, \binits{X.}},
\bauthor{\bsnm{Zhu}, \binits{L.}},
\bauthor{\bsnm{Xu}, \binits{X.}},
\bauthor{\bsnm{Xu}, \binits{Y.}},
\bauthor{\bsnm{Wang}, \binits{W.}},
\bauthor{\bsnm{Deng}, \binits{Z.}},
\bauthor{\bsnm{Heng}, \binits{P.}}:
\batitle{Deep texture-aware features for camouflaged object detection}.
\bjtitle{{IEEE} Trans. Circuits Syst. Video Technol.}
\bvolume{33}(\bissue{3}),
\bfpage{1157}--\blpage{1167}
(\byear{2023})
\doiurl{10.1109/TCSVT.2021.3126591}
\end{barticle}
\endbibitem

\bibitem[\protect\citeauthoryear{Zhu et~al.}{2021}]{multitask4}
\begin{bchapter}
\bauthor{\bsnm{Zhu}, \binits{J.}},
\bauthor{\bsnm{Zhang}, \binits{X.}},
\bauthor{\bsnm{Zhang}, \binits{S.}},
\bauthor{\bsnm{Liu}, \binits{J.}}:
\bctitle{Inferring camouflaged objects by texture-aware interactive guidance network}.
In: \bbtitle{Proceedings of the AAAI Conference on Artificial Intelligence},
vol. \bseriesno{35},
pp. \bfpage{3599}--\blpage{3607}
(\byear{2021})
\end{bchapter}
\endbibitem

\bibitem[\protect\citeauthoryear{Zhai et~al.}{2023}]{multitask5}
\begin{barticle}
\bauthor{\bsnm{Zhai}, \binits{Q.}},
\bauthor{\bsnm{Li}, \binits{X.}},
\bauthor{\bsnm{Yang}, \binits{F.}},
\bauthor{\bsnm{Jiao}, \binits{Z.}},
\bauthor{\bsnm{Luo}, \binits{P.}},
\bauthor{\bsnm{Cheng}, \binits{H.}},
\bauthor{\bsnm{Liu}, \binits{Z.}}:
\batitle{{MGL:} mutual graph learning for camouflaged object detection}.
\bjtitle{{IEEE} Trans. Image Process.}
\bvolume{32},
\bfpage{1897}--\blpage{1910}
(\byear{2023})
\doiurl{10.1109/TIP.2022.3223216}
\end{barticle}
\endbibitem

\bibitem[\protect\citeauthoryear{Liu et~al.}{2021}]{confidence}
\begin{botherref}
\oauthor{\bsnm{Liu}, \binits{J.}},
\oauthor{\bsnm{Zhang}, \binits{J.}},
\oauthor{\bsnm{Barnes}, \binits{N.}}:
Confidence-aware learning for camouflaged object detection.
CoRR
\textbf{abs/2106.11641}
(2021)
{\href{https://arxiv.org/abs/2106.11641}{{2106.11641}}}
\end{botherref}
\endbibitem

\bibitem[\protect\citeauthoryear{Zhang et~al.}{2023}]{transformer}
\begin{barticle}
\bauthor{\bsnm{Zhang}, \binits{Q.}},
\bauthor{\bsnm{Ge}, \binits{Y.}},
\bauthor{\bsnm{Zhang}, \binits{C.}},
\bauthor{\bsnm{Bi}, \binits{H.}}:
\batitle{Tprnet: camouflaged object detection via transformer-induced progressive refinement network}.
\bjtitle{Vis. Comput.}
\bvolume{39}(\bissue{10}),
\bfpage{4593}--\blpage{4607}
(\byear{2023})
\doiurl{10.1007/S00371-022-02611-1}
\end{barticle}
\endbibitem

\bibitem[\protect\citeauthoryear{Liu et~al.}{2022}]{swinv2}
\begin{bchapter}
\bauthor{\bsnm{Liu}, \binits{Z.}},
\bauthor{\bsnm{Hu}, \binits{H.}},
\bauthor{\bsnm{Lin}, \binits{Y.}},
\bauthor{\bsnm{Yao}, \binits{Z.}},
\bauthor{\bsnm{Xie}, \binits{Z.}},
\bauthor{\bsnm{Wei}, \binits{Y.}},
\bauthor{\bsnm{Ning}, \binits{J.}},
\bauthor{\bsnm{Cao}, \binits{Y.}},
\bauthor{\bsnm{Zhang}, \binits{Z.}},
\bauthor{\bsnm{Dong}, \binits{L.}}, \betal:
\bctitle{Swin transformer v2: Scaling up capacity and resolution}.
In: \bbtitle{Proceedings of the IEEE/CVF Conference on Computer Vision and Pattern Recognition},
pp. \bfpage{12009}--\blpage{12019}
(\byear{2022})
\end{bchapter}
\endbibitem

\bibitem[\protect\citeauthoryear{Wang et~al.}{2022}]{pvtv2}
\begin{barticle}
\bauthor{\bsnm{Wang}, \binits{W.}},
\bauthor{\bsnm{Xie}, \binits{E.}},
\bauthor{\bsnm{Li}, \binits{X.}},
\bauthor{\bsnm{Fan}, \binits{D.-P.}},
\bauthor{\bsnm{Song}, \binits{K.}},
\bauthor{\bsnm{Liang}, \binits{D.}},
\bauthor{\bsnm{Lu}, \binits{T.}},
\bauthor{\bsnm{Luo}, \binits{P.}},
\bauthor{\bsnm{Shao}, \binits{L.}}:
\batitle{Pvt v2: Improved baselines with pyramid vision transformer}.
\bjtitle{Computational Visual Media}
\bvolume{8}(\bissue{3}),
\bfpage{415}--\blpage{424}
(\byear{2022})
\doiurl{10.1007/s41095-022-0274-8}
\end{barticle}
\endbibitem

\bibitem[\protect\citeauthoryear{Han et~al.}{2020}]{ghostnet}
\begin{bchapter}
\bauthor{\bsnm{Han}, \binits{K.}},
\bauthor{\bsnm{Wang}, \binits{Y.}},
\bauthor{\bsnm{Tian}, \binits{Q.}},
\bauthor{\bsnm{Guo}, \binits{J.}},
\bauthor{\bsnm{Xu}, \binits{C.}},
\bauthor{\bsnm{Xu}, \binits{C.}}:
\bctitle{Ghostnet: More features from cheap operations}.
In: \bbtitle{Proceedings of the IEEE/CVF Conference on Computer Vision and Pattern Recognition},
pp. \bfpage{1580}--\blpage{1589}
(\byear{2020})
\end{bchapter}
\endbibitem

\bibitem[\protect\citeauthoryear{Vaswani et~al.}{2017}]{sa}
\begin{bchapter}
\bauthor{\bsnm{Vaswani}, \binits{A.}},
\bauthor{\bsnm{Shazeer}, \binits{N.}},
\bauthor{\bsnm{Parmar}, \binits{N.}},
\bauthor{\bsnm{Uszkoreit}, \binits{J.}},
\bauthor{\bsnm{Jones}, \binits{L.}},
\bauthor{\bsnm{Gomez}, \binits{A.N.}},
\bauthor{\bsnm{Kaiser}, \binits{L.}},
\bauthor{\bsnm{Polosukhin}, \binits{I.}}:
\bctitle{Attention is all you need}.
In: \bbtitle{Proceedings of the 31st International Conference on Neural Information Processing Systems}.
\bsertitle{NIPS'17},
pp. \bfpage{6000}--\blpage{6010}.
\bpublisher{Curran Associates Inc.},
\blocation{Red Hook, NY, USA}
(\byear{2017})
\end{bchapter}
\endbibitem

\bibitem[\protect\citeauthoryear{Watters et~al.}{2019}]{sbd}
\begin{botherref}
\oauthor{\bsnm{Watters}, \binits{N.}},
\oauthor{\bsnm{Matthey}, \binits{L.}},
\oauthor{\bsnm{Burgess}, \binits{C.P.}},
\oauthor{\bsnm{Lerchner}, \binits{A.}}:
Spatial broadcast decoder: {A} simple architecture for learning disentangled representations in vaes.
CoRR
\textbf{abs/1901.07017}
(2019)
{\href{https://arxiv.org/abs/1901.07017}{{1901.07017}}}
\end{botherref}
\endbibitem

\bibitem[\protect\citeauthoryear{Le et~al.}{2019}]{camo}
\begin{barticle}
\bauthor{\bsnm{Le}, \binits{T.-N.}},
\bauthor{\bsnm{Nguyen}, \binits{T.V.}},
\bauthor{\bsnm{Nie}, \binits{Z.}},
\bauthor{\bsnm{Tran}, \binits{M.-T.}},
\bauthor{\bsnm{Sugimoto}, \binits{A.}}:
\batitle{Anabranch network for camouflaged object segmentation}.
\bjtitle{Computer Vision and Image Understanding}
\bvolume{184},
\bfpage{45}--\blpage{56}
(\byear{2019})
\doiurl{10.1016/j.cviu.2019.04.006}
\end{barticle}
\endbibitem

\bibitem[\protect\citeauthoryear{Skurowski et~al.}{2018}]{chameleon}
\begin{botherref}
\oauthor{\bsnm{Skurowski}, \binits{P.}},
\oauthor{\bsnm{Abdulameer}, \binits{H.}},
\oauthor{\bsnm{Błaszczyk}, \binits{J.}},
\oauthor{\bsnm{Depta}, \binits{T.}},
\oauthor{\bsnm{Kornacki}, \binits{A.}},
\oauthor{\bsnm{Kozieł}, \binits{P.}}:
Animal camouflage analysis: Chameleon database.
Unpublished Manuscript
(2018)
\end{botherref}
\endbibitem

\bibitem[\protect\citeauthoryear{Song et~al.}{2023}]{fsnet}
\begin{barticle}
\bauthor{\bsnm{Song}, \binits{Z.}},
\bauthor{\bsnm{Kang}, \binits{X.}},
\bauthor{\bsnm{Wei}, \binits{X.}},
\bauthor{\bsnm{Liu}, \binits{H.}},
\bauthor{\bsnm{Dian}, \binits{R.}},
\bauthor{\bsnm{Li}, \binits{S.}}:
\batitle{Fsnet: Focus scanning network for camouflaged object detection}.
\bjtitle{IEEE Transactions on Image Processing}
\bvolume{32},
\bfpage{2267}--\blpage{2278}
(\byear{2023})
\doiurl{10.1109/TIP.2023.3266659}
\end{barticle}
\endbibitem

\bibitem[\protect\citeauthoryear{Wang et~al.}{2018}]{scrn}
\begin{barticle}
\bauthor{\bsnm{Wang}, \binits{X.}},
\bauthor{\bsnm{Ma}, \binits{H.}},
\bauthor{\bsnm{Chen}, \binits{X.}},
\bauthor{\bsnm{You}, \binits{S.}}:
\batitle{Edge preserving and multi-scale contextual neural network for salient object detection}.
\bjtitle{IEEE Transactions on Image Processing}
\bvolume{27}(\bissue{1}),
\bfpage{121}--\blpage{134}
(\byear{2018})
\doiurl{10.1109/tip.2017.2756825}
\end{barticle}
\endbibitem

\bibitem[\protect\citeauthoryear{Gao et~al.}{2020}]{csnet}
\begin{bbook}
\bauthor{\bsnm{Gao}, \binits{S.-H.}},
\bauthor{\bsnm{Tan}, \binits{Y.-Q.}},
\bauthor{\bsnm{Cheng}, \binits{M.-M.}},
\bauthor{\bsnm{Lu}, \binits{C.}},
\bauthor{\bsnm{Chen}, \binits{Y.}},
\bauthor{\bsnm{Yan}, \binits{S.}}:
\bbtitle{Highly efficient salient object detection with 100k parameters},
pp. \bfpage{702}--\blpage{721}
(\byear{2020}).
\bcomment{Springer}
\end{bbook}
\endbibitem

\bibitem[\protect\citeauthoryear{Zhao et~al.}{2019}]{egnet}
\begin{bchapter}
\bauthor{\bsnm{Zhao}, \binits{J.-X.}},
\bauthor{\bsnm{Liu}, \binits{J.-J.}},
\bauthor{\bsnm{Fan}, \binits{D.-P.}},
\bauthor{\bsnm{Cao}, \binits{Y.}},
\bauthor{\bsnm{Yang}, \binits{J.}},
\bauthor{\bsnm{Cheng}, \binits{M.-M.}}:
\bctitle{Egnet: Edge guidance network for salient object detection}.
In: \bbtitle{Proceedings of the IEEE/CVF International Conference on Computer Vision},
pp. \bfpage{8779}--\blpage{8788}
(\byear{2019})
\end{bchapter}
\endbibitem

\bibitem[\protect\citeauthoryear{Li et~al.}{2021}]{ujsc}
\begin{bchapter}
\bauthor{\bsnm{Li}, \binits{A.}},
\bauthor{\bsnm{Zhang}, \binits{J.}},
\bauthor{\bsnm{Lv}, \binits{Y.}},
\bauthor{\bsnm{Liu}, \binits{B.}},
\bauthor{\bsnm{Zhang}, \binits{T.}},
\bauthor{\bsnm{Dai}, \binits{Y.}}:
\bctitle{Uncertainty-aware joint salient object and camouflaged object detection}.
In: \bbtitle{Proceedings of the IEEE/CVF Conference on Computer Vision and Pattern Recognition},
pp. \bfpage{10071}--\blpage{10081}
(\byear{2021})
\end{bchapter}
\endbibitem

\bibitem[\protect\citeauthoryear{Zhai et~al.}{2023}]{mgl}
\begin{barticle}
\bauthor{\bsnm{Zhai}, \binits{Q.}},
\bauthor{\bsnm{Li}, \binits{X.}},
\bauthor{\bsnm{Yang}, \binits{F.}},
\bauthor{\bsnm{Jiao}, \binits{Z.}},
\bauthor{\bsnm{Luo}, \binits{P.}},
\bauthor{\bsnm{Cheng}, \binits{H.}},
\bauthor{\bsnm{Liu}, \binits{Z.}}:
\batitle{Mgl: Mutual graph learning for camouflaged object detection}.
\bjtitle{IEEE Transactions on Image Processing}
\bvolume{32},
\bfpage{1897}--\blpage{1910}
(\byear{2023})
\doiurl{10.1109/tip.2022.3223216}
\end{barticle}
\endbibitem

\bibitem[\protect\citeauthoryear{Mei et~al.}{2021}]{pfnet}
\begin{bchapter}
\bauthor{\bsnm{Mei}, \binits{H.}},
\bauthor{\bsnm{Ji}, \binits{G.-P.}},
\bauthor{\bsnm{Wei}, \binits{Z.}},
\bauthor{\bsnm{Yang}, \binits{X.}},
\bauthor{\bsnm{Wei}, \binits{X.}},
\bauthor{\bsnm{Fan}, \binits{D.-P.}}:
\bctitle{Camouflaged object segmentation with distraction mining},
pp. \bfpage{8768}--\blpage{8777}
(\byear{2021}).
\doiurl{10.1109/CVPR46437.2021.00866}
\end{bchapter}
\endbibitem

\bibitem[\protect\citeauthoryear{Fan et~al.}{2020}]{pranet}
\begin{bchapter}
\bauthor{\bsnm{Fan}, \binits{D.-P.}},
\bauthor{\bsnm{Ji}, \binits{G.-P.}},
\bauthor{\bsnm{Zhou}, \binits{T.}},
\bauthor{\bsnm{Chen}, \binits{G.}},
\bauthor{\bsnm{Fu}, \binits{H.}},
\bauthor{\bsnm{Shen}, \binits{J.}},
\bauthor{\bsnm{Shao}, \binits{L.}}:
\bctitle{Pranet: Parallel reverse attention network for polyp segmentation}.
In: \bbtitle{International Conference on Medical Image Computing and Computer-assisted Intervention},
pp. \bfpage{263}--\blpage{273}
(\byear{2020}).
\bcomment{Springer}
\end{bchapter}
\endbibitem

\bibitem[\protect\citeauthoryear{Jiang et~al.}{2023}]{jcnet}
\begin{barticle}
\bauthor{\bsnm{Jiang}, \binits{X.}},
\bauthor{\bsnm{Cai}, \binits{W.}},
\bauthor{\bsnm{Ding}, \binits{Y.}},
\bauthor{\bsnm{Wang}, \binits{X.}},
\bauthor{\bsnm{Hong}, \binits{D.}},
\bauthor{\bsnm{Yang}, \binits{Z.}},
\bauthor{\bsnm{Gao}, \binits{W.}}:
\batitle{Camouflaged object segmentation based on joint salient object for contrastive learning}.
\bjtitle{IEEE Transactions on Instrumentation and Measurement}
\bvolume{72},
\bfpage{1}--\blpage{16}
(\byear{2023})
\doiurl{10.1109/TIM.2023.3306520}
\end{barticle}
\endbibitem

\bibitem[\protect\citeauthoryear{Yang et~al.}{2018}]{denseaspp}
\begin{bchapter}
\bauthor{\bsnm{Yang}, \binits{M.}},
\bauthor{\bsnm{Yu}, \binits{K.}},
\bauthor{\bsnm{Zhang}, \binits{C.}},
\bauthor{\bsnm{Li}, \binits{Z.}},
\bauthor{\bsnm{Yang}, \binits{K.}}:
\bctitle{Denseaspp for semantic segmentation in street scenes}.
In: \bbtitle{Proceedings of the IEEE Conference on Computer Vision and Pattern Recognition},
pp. \bfpage{3684}--\blpage{3692}
(\byear{2018})
\end{bchapter}
\endbibitem

\end{thebibliography}

\end{document}